DreamNLP: Novel NLP System for Clinical Report Metadata Extraction using Count Sketch Data Streaming Algorithm: Preliminary Results

Sanghyun Choi[1], Nikita Ivkin[1], Vladimir Braverman[1], Michael A. Jacobs[2,3]

[1]Department of Computer Science, Johns Hopkins University, Baltimore, MD 21218, USA

[2]The Russell H. Morgan Department of Radiology and Radiological Sciences,

Division of Cancer Imaging and [3]Sidney Kimmel Comprehensive Cancer Center

Johns Hopkins University School of Medicine, Baltimore, MD 21205, USA

**Corresponding Author**
Michael A. Jacobs
Traylor Blg, Rm 309,
The Russell H. Morgan Department of Radiology and Radiological Sciences,
The Johns Hopkins University School of Medicine,
712 Rutland Ave. Baltimore MD 21205 USA
Phone: 410-955-7483
Email: mikej@mri.jhu.edu

## ABSTRACT

Extracting information from electronic health records (EHR) is a challenging task since it requires prior knowledge of the reports and some natural language processing algorithm (NLP). With the growing number of EHR implementations, such knowledge is increasingly challenging to obtain in an efficient manner. We address this challenge by proposing a novel methodology to analyze large sets of EHRs using a modified Count Sketch data streaming algorithm termed DreamNLP. By using DreamNLP, we generate a dictionary of frequently occurring terms or heavy hitters in the EHRs using low computational memory compared to conventional counting approach other NLP programs use. We demonstrate the extraction of the most important breast diagnosis features from the EHRs in a set of patients that underwent breast imaging. Based on the analysis, extraction of these terms would be useful for defining important features for downstream tasks such as machine learning for precision medicine.

## INTRODUCTION

Clinical reports often contain detailed snapshots of a patients' health that could be used for studying the "hidden" connections to a particular disease. With the advent of electronic health record (EHR) system and sophisticated Natural Language Processing (NLP) techniques, the ability to extract various information from the EHR have become an active areas of research[2]. For example, whether a heart failure or a genetic risk for diabetes can be identified[3,4]. Therefore, EHRs play a major role as useful source of important medical data to clinicians. However, to extract and collect data or terms across many different medical records is a difficult task. There are NLP tools such as Apache cTAKES to aid in information extraction, but these methods can be computationally and memory intensive[5]. Moreover, some NLP programs only provide extracted terms and it may be difficult to build quantitative relationships between these terms and what is in the report.

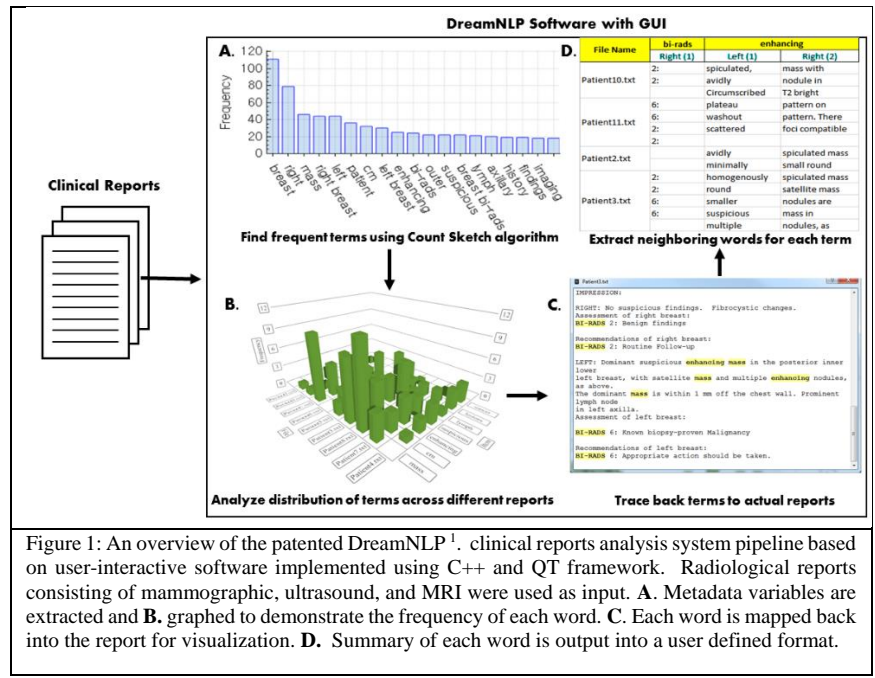

Figure 1: An overview of the patented DreamNLP [1]. clinical reports analysis system pipeline based on user-interactive software implemented using C++ and QT framework. Radiological reports consisting of mammographic, ultrasound, and MRI were used as input. **A**. Metadata variables are extracted and **B**. graphed to demonstrate the frequency of each word. **C**. Each word is mapped back into the report for visualization. **D**. Summary of each word is output into a user defined format.

In addition, it is difficult to know which set of data could potentially be obtained from the reports without a strong prior knowledge of their contents. One of the simplest ways to gain a basic understanding of large corpora of text is to observe and extract the frequently occurring terms within the reports. For example, word clouds, consisting of frequent terms, are often effective in visually summarizing particular texts[6]. Yet, finding frequent terms from large corpora is a challenging task by itself since every unique term in the data has to be stored in memory if approached in a naïve way.

In order to address these issues, we propose a novel methodology in which we first find a set of frequent terms among the clinical reports to form a dictionary of terms that could serve as basis sets for information extraction strategy. This is achieved in a memory efficient way by applying the Count Sketch streaming algorithm[7,8]. Several studies have indicated the robustness of sketch algorithms in the context of large scale NLP tasks[9-11]. We extend this and use Shannon entropy to quantitatively evaluate how well the terms in the dictionary are distributed across all the reports. To better assess whether a term is associated with extractable data, we trace it back to the reports and observe its context and structure. The overall pipeline for DreamNLP is shown in Figure 1.

We make the following main contributions in this paper:

- To the best of our knowledge, this is the first application of Count Sketch algorithm to clinical reports to enable large scale data analysis.
- We present DreamNLP (**D**ata St**ream**ing **NLP**), a methodical system to extract, analyze and annotate clinical reports[1].
- We demonstrate the robustness of the DreamNLP system for extracting known features defined by structured radiological reporting for improved diagnosis.
- We present a new multidimensional visualization tool (Figure 2) to determine the relationships between each object (i.e. terms, reports, frequencies).

**METHODS**

**Overview**

Theoretically, clinical reports (or any report) can be viewed as a data stream of terms (in this paper, a "term" could mean a single word or a sequential set of words). We use this concept to initiate the input into our software platform. Initially, the analysis begins with an input of clinical reports extracted from our electronic database. Then, DreamNLP finds top-$k$ most frequent terms (i.e. heavy hitters) using the Count Sketch streaming algorithm with very low memory usage. Upon obtaining the heavy hitters, they are evaluated using Shannon entropy, a metric that gives us an idea of whether the majority of reports contain a particular term. This gives an indication of the strength of the term or terms. We have developed graphic user interface for a visualization tool for viewing how the terms are actually distributed and used in the reports so that they can be better assessed for anticipated downstream tasks. Finally, we generate an output of the neighboring words for each term in a user defined format, which could be used for future post-processing or data analysis.

**Finding frequent terms using a streaming algorithm**

The basic Count Sketch algorithm (CSA) solves the problem of finding top-$k$ most frequent items in data stream[7,12]. Compared to naïve approach to this problem which implies keeping track of each item's frequency, CSA requires relatively low memory, therefore has lower space complexity than other NLP algorithms. CSA's approximation level and probability of success can be controlled via input parameters.

As shown in Figure 3, the "Count Sketch" algorithm is simply a $t \times b$ matrix $C$ of counters that can be viewed as $t$ hash tables with $b$ buckets for each hash table.

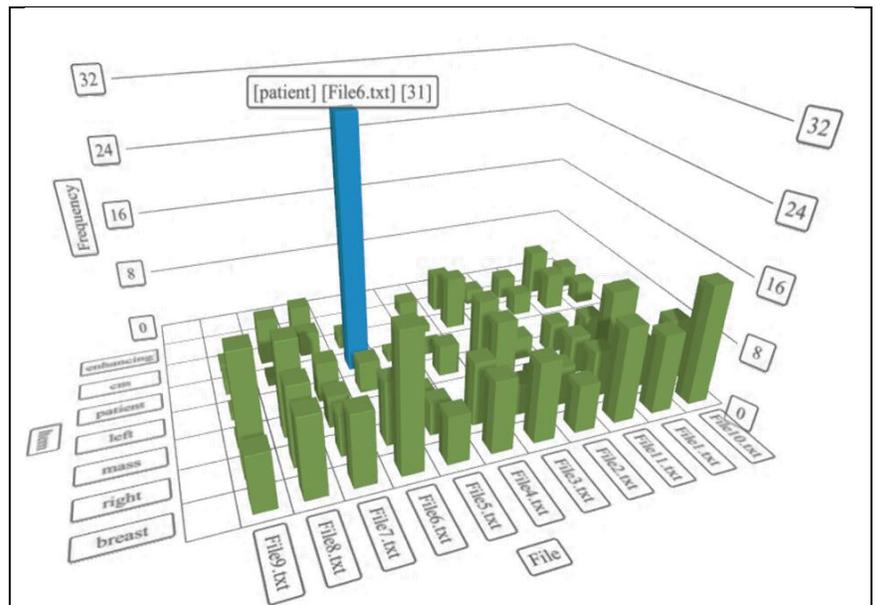

Figure 2: A multidimensional histogram showing the relationship between frequencies, terms, and reports derived from DreamNLP, where one or several reports can be visualized together for comparison to other records or objects for improved diagnostics.

For a particular choice of $k$, the parameters $t$ and $b$ play an important role in determining the accuracy of estimated frequencies. Now, suppose we have a stream of items $\{o_1,...,o_n\}$, where $n$ is the total number of items in the data stream. Also, let $\{h_1,...,h_t\}$ and $\{s_1,...s_t\}$ be two sets of $t$ hash functions that map $o_l$ to a particular value in $\{1,...,b\}$ and $\{-1,1\}$, respectively, where $o_l$ is some $l$-th item in the stream. The Count Sketch algorithm involves two operations:

- Updating counters:

    For each $i$-th row in matrix $C$, update counter in $h_i(o_l)$-th column by adding $s_i(o_l)$.

- Estimating frequency:

    For each $i$-th row in matrix $C$, compute $C_{ij} \cdot s_i(o_l)$, where $j=h_i(o_l)$ as above, and estimate the frequency of $o_l$ by their median.

Based on these two operations, the Count Sketch algorithm outputs an estimate of the frequencies for the top $k$ items. In order to make Count Sketch applicable to our context (i.e. stream of terms of clinical reports), we utilize the MurmurHash, a robust string hashing algorithm in the design of hash functions $h$ and $s$[13,14].

**Evaluation of terms using Shannon entropy**

We evaluate the significance of each frequent term by observing how these terms are distributed across different reports. We quantitatively examine the distribution by using the Shannon entropy, which is denoted by $H$, and defined as follows[15]:

$$H = -\sum_{i=1}^{n}(p_i \log p_i)$$

Here $n$ is the number of possible events and $p_i$ is the probability that $i$-th event occurs. If $H$ attains its maximum value, it implies that all events are equally likely (i.e. $p_i = 1/n$ for all $i$). In our context, the $i$-th event corresponds to a particular term appearing in the $i$-th report when there are $n$ reports. Therefore, a high value of $H_T$, the Shannon entropy for a particular term $T$, implies that $T$ appears in most of the reports while a low value implies that $T$ only appears in a small subset of reports. Based on this property, $H_T$ can be used as a measure for $T$'s usefulness, depending on the anticipated downstream task. The value $H_T$ is obtained by computing the values of $p_i$ as "$n_i / n_{total}$" where $n_i$ is the number of occurrences of $T$ in the $i$-th report and $n_{total}$ is the total number of occurrences of $T$ in the entire set of reports. Since there are at most $k$ terms under consideration we can now compute exact values of $n_i$.

**Context analysis of terms**

The final steps of the DreamNLP pipeline involve better understanding the context of each term. To that end, our visualization tool enables viewing of the terms use in the actual reports. Moreover, we can generate many different outputs for further data analysis or applications. For example, the number of words to the left and right of a term can be configured to extract information associated with string terms. Finally, we have developed a multidimensional histogram to visualize and determine the relationship between frequencies, terms, and reports.

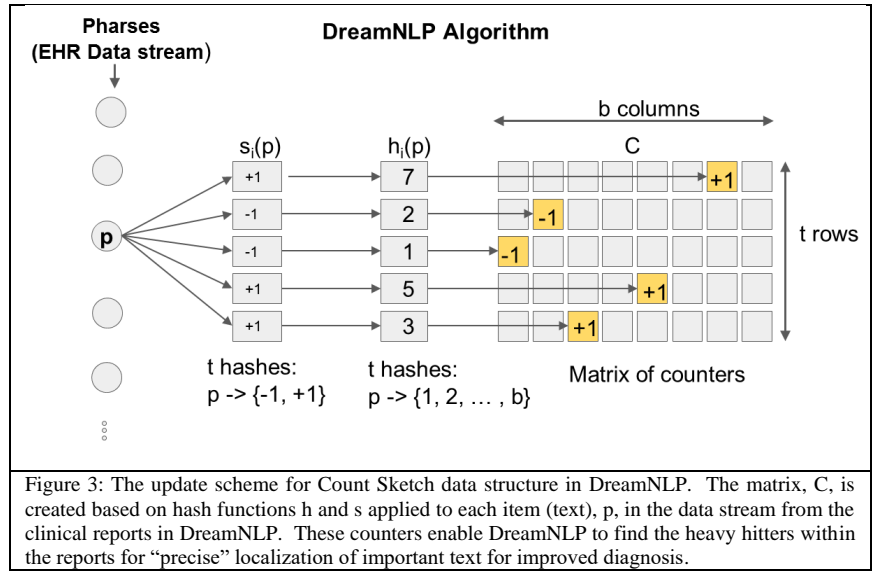

Figure 3: The update scheme for Count Sketch data structure in DreamNLP. The matrix, C, is created based on hash functions h and s applied to each item (text), p, in the data stream from the clinical reports in DreamNLP. These counters enable DreamNLP to find the heavy hitters within the reports for "precise" localization of important text for improved diagnosis.

**Clinical database used for the DreamNLP**

Clinical reports of clinical encounters, pathology and imaging were obtained from patients who underwent breast imaging consisting of mammography, ultrasound, and MRI[16]. For this study, all reports were anonymized with no patient health identifiers within the report. The reports were obtained by an approved IRB study following HIPPA criteria. Breast imaging reports are structured and must adhere to the Breast Imaging Reporting and Data System (BI-RADS) format[17,18]. Briefly, the BI-RADS scoring gives a numerical index to the degree of potential malignancy of breast tumors. The BI-RADS scale ranges from 1 to 6, where 1-3 is considered normal, 4-5 is considered highly suspicious and 6 is proven malignancy. Moreover, each breast is described by defined features, such as breast density, lesion shape, size, and distribution. Each clinical report was converted to a text file and read into DreamNLP.

**RESULTS**

DreamNLP was able to distinguish between imaging and non-imaging clinical reports and filter out non-BI-RADS reports (a small number relative to BI-RADS reports) by using low Shannon entropy such as "FDG uptake" and the reports that were missing the high entropy terms such as "BI-RADS", morphology, and "shape". DreamNLP selected and analyzed 46 records (75%) out of the 61 reports available with BI-RADS reporting.

DreamNLP was applied to 24,487 terms in the selected reports to extract the top 100 terms and it took about 20 milliseconds. The parameters $t$ and $b$ were chosen such that the results closely matched true values (i.e. $t = 20$ and $b = 10,000$). Careful analysis of the 100 terms revealed useful information demonstrated that DreamNLP was able to extract specific features from the BI-RADS reports. Table 1 summarizes the analysis of the significant terms based on the entropy evaluation and context analysis.

Table 1: DreamNLP text features extracted from the electronic health record.

| Dictionary Generation | | Evaluation | Context Analysis | Insight |
|---|---|---|---|---|
| Rank[1] | Term(s) | Shannon Entropy[2] | Sample Neighboring Words | Potential Feature |
| 5, 7 | right breast, left breast | 94.38, 94.42 | *RIGHT BREAST:* <br> *LEFT BREAST:* | Left/Right Distinction |
| 6 | cm | 89.33 | mass measuring 1.1 x 0.8 x 0.9 *cm* | Size of Lesion |
| 8 | mass | 83.65 | lobulated, spiculated *mass* | Type of Lesion |
| 13, 20 | bi-rads, assessment | 84.69, 96.19 | *BI-RADS* 5: Highly suspicious … <br> Assessment: 6 Known malignancy | BI-RADS Scores |
| 41, 95 | o'clock, quadrant | 72.39, 70.58 | at the two *o'clock* position <br> right upper outer *quadrant* | Site of Lesion |
| 34 | history | 92.81 | strong family *history* of breast cancer | Family history |
| 54, 74 | density, dense | 96.11, 88.34 | Breast *density*: Moderately *dense*. | Breast Density |

1: The rank of a term based on its number of occurrences in the entire set of reports (estimated by Count Sketch algorithm) 2: Percentage with respect to maximum possible value

The high entropy values associated with each term indicate that these terms are more or less present in the majority of reports and thus can be used for information extraction and feature generation. For example, we extracted "BI-RADS Scores" from each report using the terms "BI-RADS", "Left or Right Breast", and "mass measuring", thus enabling the classification of patients into malignant or benign groups.

**CONCLUSION**

We have demonstrated that DreamNLP can extract significant features related to breast diagnosis from radiological clinical reports using key BI-RADS features for improved diagnostic performance. In addition, DreamNLP uses very low computational memory and minimal prior knowledge of the structure of reports with high accuracy of correct extraction of terms. These selected terms then could be used for populating an "in house" corpora from local populations for further informatics work. Besides the possibility of obtaining potential features, finding frequent terms can also reveal other useful insights from reports. For example, considering that most of the reports analyzed were malignant cases, the fact that the term "upper outer" appeared among the top 100 terms provides evidence that the upper outer quadrant of the breast is frequently vulnerable to carcinoma[19]. These finding can be exploited to discover "hidden" connections between different variables or modalities and be integrated into the "big data" approach.

One potential limitation of DreamNLP for frequent terms analysis is that it might miss terms which appear consistently throughout many documents but with relatively less frequency than other terms, such as "age" in the case of BI-RADS. However, this limitation could be controlled to a certain extent by increasing the number of the top-$k$ terms to find and filtering out more common words, such as, "be, to, as", and linking verbs.

The DreamNLP methodology is flexible for other extensions and richer seamless analysis. For example, applications can be incorporated in the software using negation by applying NegEx algorithm to some of the terms[20]. We are investigating how negated expressions could be used as features for machine learning and quality assurance monitoring. Finally, our methodology can be generalized to analyze other types of electronic documents besides clinical reports, for example, legal reports.

**ACKNOWLEDGMENTS:**

Funding from NIH grants 5P30CA006973 (IRAT), R01CA190299, U01CA140204 and the K40 GPU donation from the Nvidia Corporation.

**Author Statement**

MAJ, SC, NI, and VB developed the model. SG collected the data. MAJ and SC. analyzed the data. SC, NI, VB, and MAJ did the study design. NI and VB. assisted in evaluation of the Streaming Algorithm model of the study. SC, NI, and MAJ wrote the manuscript. All authors contributed to the review and revisions of the manuscript. All the authors have seen and approved the final version of the manuscript.

**FIGURES**

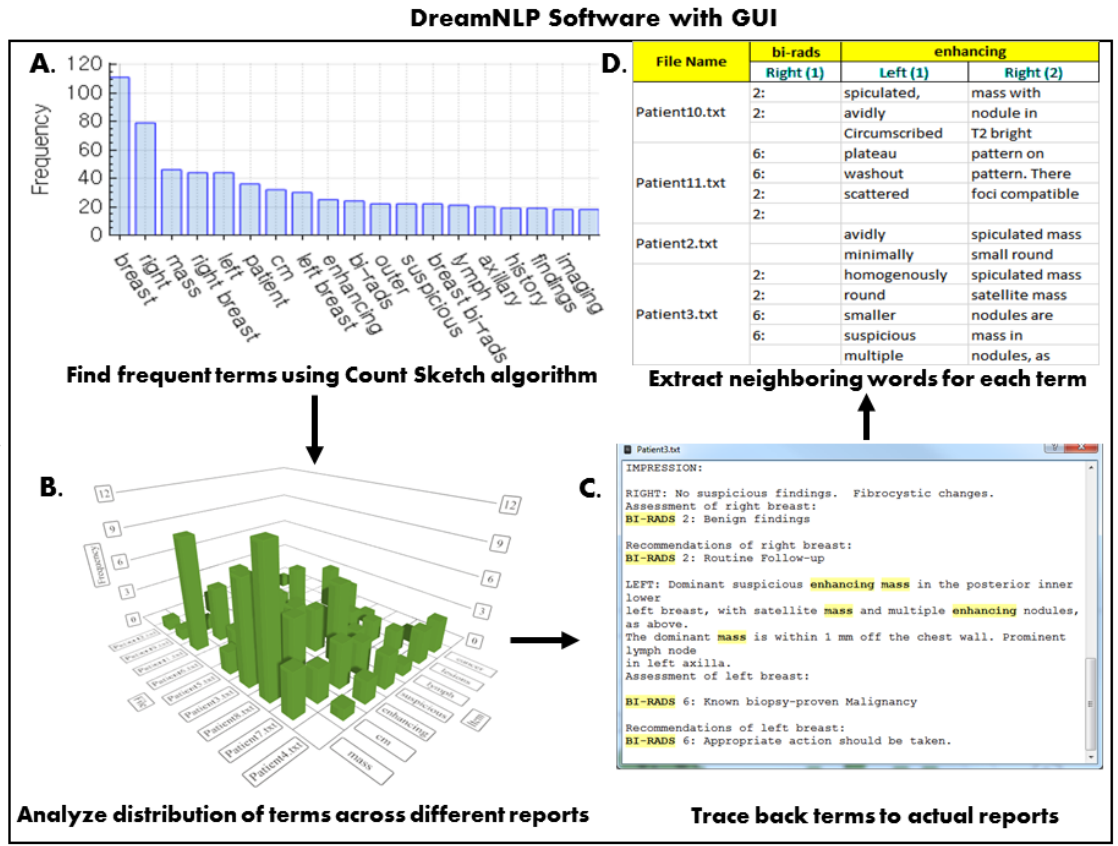

Figure 1: An overview of the DreamNLP clinical reports analysis system pipeline based on user-interactive software implemented using C++ and QT framework. Radiological reports consisting of mammographic, ultrasound, and MRI were used as input. A. Metadata variables are extracted and B. graphed to demonstrate the frequency of each word. C. Each word is mapped back into the report for visualization. D. Summary of each word is output into a user defined format.

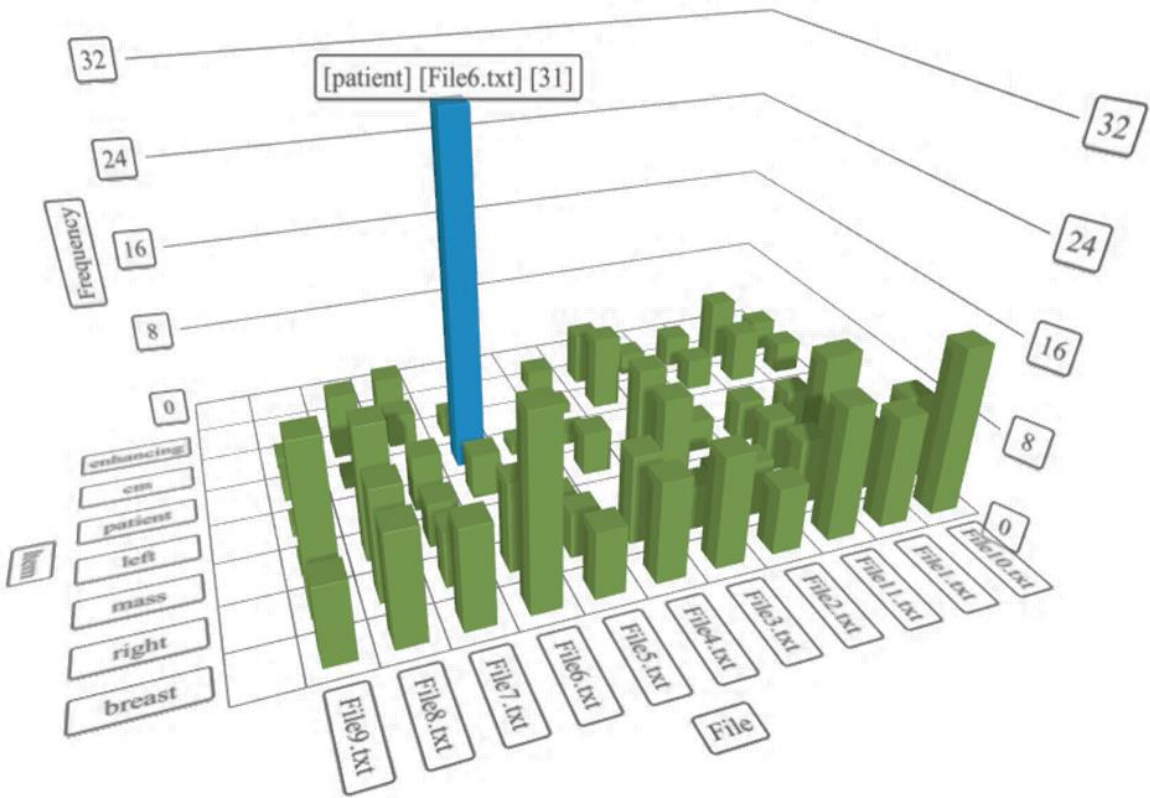

Figure 2: A multidimensional histogram showing the relationship between frequencies, terms, and reports derived from DreamNLP, where one or many reports can be visualized together for comparison to other records for improved diagnostics.

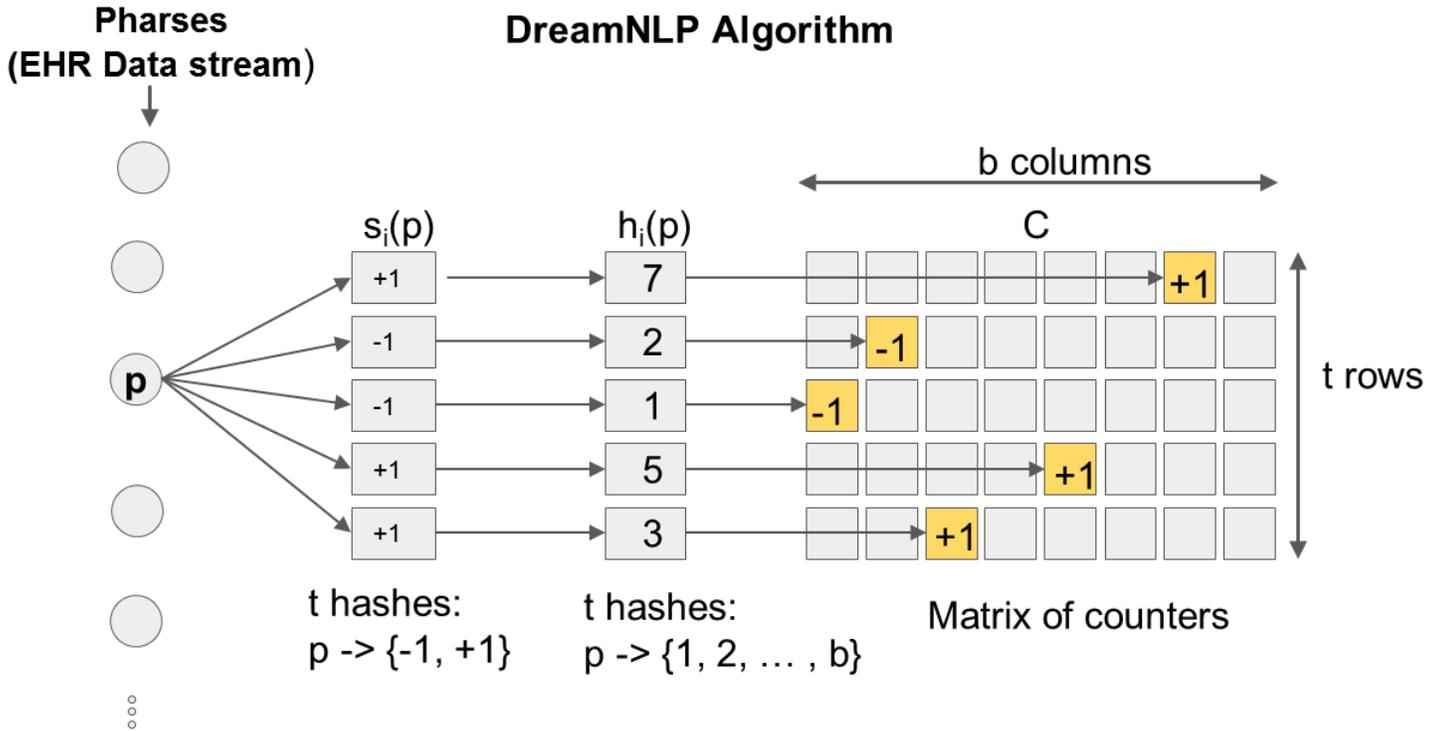

Figure 3: The update scheme for Count Sketch data structure in DreamNLP. The matrix, *C*, is created based on hash functions *h* and *s* applied to each item (text), *p*, in the data stream from the clinical reports in DreamNLP. These counters enable DreamNLP to find the heavy hitters within the reports for "precise" localization of important text for improved diagnosis.